\documentclass[sigconf]{acmart}
\AtBeginDocument{%
  \providecommand\BibTeX{{%
    \normalfont B\kern-0.5em{\scshape i\kern-0.25em b}\kern-0.8em\TeX}}}

\copyrightyear{2022} 
\acmYear{2022} 
\setcopyright{acmcopyright}
\acmConference[MM '22]{Proceedings of the 30th ACM
International Conference on Multimedia}{October 10--14, 2022}{Lisboa, Portugal}
\acmBooktitle{Proceedings of the 30th ACM International Conference on Multimedia
(MM '22), October 10--14, 2022, Lisboa, Portugal}
\acmPrice{15.00}
\acmDOI{10.1145/3503161.3547938}
\acmISBN{978-1-4503-9203-7/22/10}

\usepackage{multirow}
\usepackage{enumitem}
\usepackage{subcaption}
\usepackage{soul}
\usepackage{amsmath} 
\usepackage{balance}
\setlist[itemize]{leftmargin=15pt}

\newcommand{\up}[1]{\textcolor{blue}{#1}}
\newcommand{\down}[1]{\textcolor{red}{#1}}
\begin{document}

\title{Task-adaptive Spatial-Temporal Video Sampler for Few-shot Action Recognition}


\author{Huabin Liu}
\affiliation{%
  \institution{Shanghai Jiao Tong University}
  \city{Shanghai}
  \country{China}}
\email{huabinliu@sjtu.edu.cn}

\author{Weixian Lv}
\affiliation{%
  \institution{Shanghai Jiao Tong University}
  \city{Shanghai}
  \country{China}}
\email{darkterror@sjtu.edu.cn}

\author{John See}
\affiliation{%
  \institution{Heriot-Watt University Malaysia}
  \city{Putrajaya}
  \country{Malaysia}}
\email{J.See@hw.ac.uk}

\author{Weiyao Lin}
\authornote{Corresponding author}
\affiliation{%
  \institution{Shanghai Jiao Tong University}
  \city{Shanghai}
  \country{China}}
\email{wylin@sjtu.edu.cn}
\renewcommand{\shortauthors}{Huabin Liu, Weixian Lv, John See, \& Weiyao Lin}

% Authors, replace the red X's with your assigned DOI string during the rightsreview eform process.
%% Your DOI link will become active when the proceedings appears in the DL.

\settopmatter{printacmref=true}
\begin{abstract}
A primary challenge faced in few-shot action recognition is inadequate video data for training. To address this issue, current methods in this field mainly focus on devising algorithms at the feature level while little attention is paid to processing input video data. Moreover, existing frame sampling strategies may omit critical action information in temporal and spatial dimensions, which further impacts video utilization efficiency.
In this paper, we propose a novel video frame sampler for few-shot action recognition to address this issue, where task-specific spatial-temporal frame sampling is achieved via a temporal selector (TS) and a spatial amplifier (SA). Specifically, our sampler first scans the whole video at a small computational cost to obtain a global perception of video frames. The TS plays its role in selecting top-$T$ frames that contribute most significantly and subsequently. The SA emphasizes the discriminative information of each frame by amplifying critical regions with the guidance of saliency maps. We further adopt task-adaptive learning to dynamically adjust the sampling strategy according to the episode task at hand. Both the implementations of TS and SA are differentiable for end-to-end optimization, facilitating seamless integration of our proposed sampler with most few-shot action recognition methods. Extensive experiments show a significant boost in the performances on various benchmarks including long-term videos. The code is available at \href{https://github.com/R00Kie-Liu/Sampler}{\textcolor{magenta}{https://github.com/R00Kie-Liu/Sampler}}.

\end{abstract}

% <concept_id>10010147.10010178.10010224</concept_id>
% <concept_desc>Computing methodologies~Computer vision</concept_desc>




\maketitle
\section{Introduction}
\label{sec:intro}
Recent years have witnessed spectacular developments in video action recognition~\cite{hieve2020, keypoint_comp2020,yufeng}. Most current approaches for action recognition employ deep learning models~\cite{c3d2015,I3d2017,TSN2016}, which are expected to achieve higher performance but require numerous labeled video data for training. Since the expensive cost of collecting and annotating video data, sometimes very few video data samples are available in real-world applications. Consequently, the \emph{few-shot} video recognition task, which aims to learn a robust action classifier with very few training samples, has attracted much attention. 
\begin{figure}[ht]
    \begin{center}
        \includegraphics[width=1\linewidth]{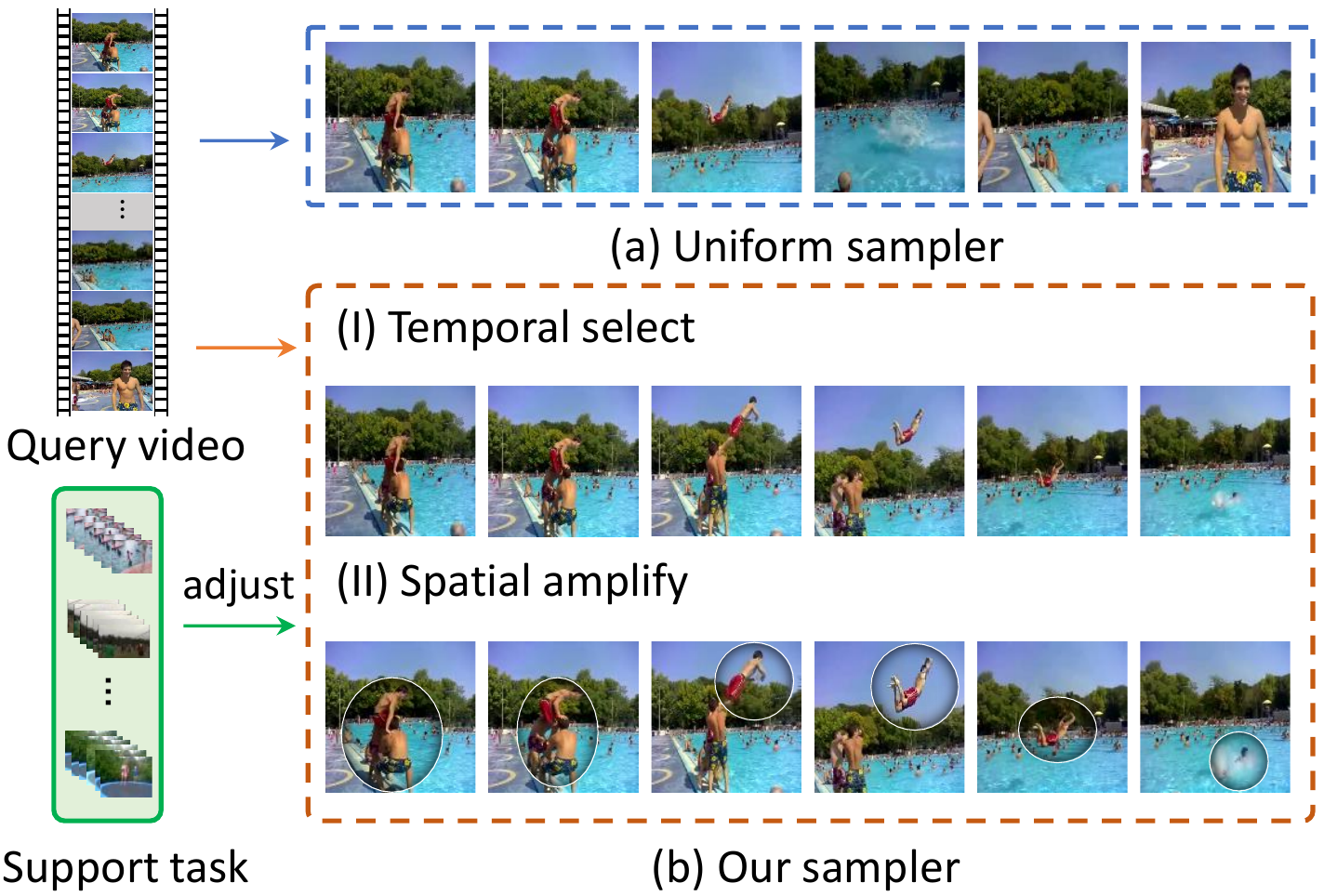}
    \end{center}
    \caption{(a) A uniform sampler may overlook frames containing key actions. Critical regions involving the actors and objects may be too small to be properly recognized. (b) Our proposed sampler is able to (I) select frames from an entire video that contribute most to few-shot recognition, (II) amplify discriminative regions in each frame. This sampling strategy is also dynamically adjusted for each video according to the episode task at hand.}
    %\vspace{-0.75em}
    \label{fig:intro}
\end{figure}

Existing few-shot action recognition methods mainly follow the metric-learning paradigm, which classifies videos by performing comparisons under specific learned metrics. Some approaches perform temporal alignment among videos to obtain consistent features for recognition~\cite{otam2020,tarn2019,ta2n2022}, while others learn to enhance video representation by various mechanisms~\cite{cmn-j2020,ARN2020,trx2021}. However, most studies in this field only focus on designing operations at the feature level, while few attempts pay close attention to the processing of input video data. Moreover, current methods adopt a uniform input sampling strategy, \emph{i.e.,} $T$ frames are sampled uniformly and the whole frame is scaled to a fixed size. This strategy may omit critical action information (as illustrated in Fig.~\ref{fig:intro}) in temporal and spatial dimensions. Thus, this further (1) reduces the efficiency in video data utilization, and (2) poses a stumbling block to optimal feature metric learning. Hence, it is necessary to explore solutions to mitigate this issue from the data source.

In this paper, we devise a novel video sampler for few-shot action recognition to solve this issue from both temporal and spatial aspects. Our sampler consists of a temporal selector (TS) and spatial amplifier (SA), which works hand-in-hand to enable task-specific spatial-temporal frame sampling for support and query videos. Concretely, we first utilize a lightweight network for scanning the whole video with a dense frame frequency; this obtains a global perception of the video. For temporal sampling, we aim to select $T$ frames that contribute most to few-shot recognition instead of naively increasing the number of input frames. The TS first evaluates each frame to obtain importance scores, which are used in the selection of $T$ frames with the highest scores. Since the top-$T$ selection operation is non-differentiable, we implement it by the perturbed maximum method~\cite{perturb2020}. Meanwhile, the SA plays its role by emphasizing the critical areas for recognition while preserving most information in the frames selected by TS. To this end, we generate a saliency map for each frame, which indicates the discriminativeness of each pixel. Guided by saliency, the SA performs 2-D inverse transform sampling over frames to amplify salient regions. Besides, since each video relies on all of the other samples in the current episode for classification, we further adopt a task-adaptive learner to dynamically adjust the sampling strategy of each video according to the current episode task at hand. Our proposed sampler can be plugged into existing few-shot action recognition methods and be trained altogether end-to-end. We equip them with our sampler and conduct comprehensive experiments on various video datasets (UCF, HMDB, SSv2, and Kinetics). Moreover, we also report few-shot results on the long-term video dataset ActivityNet~\cite{activity2015} for the first time, to our knowledge. Overall, results demonstrate that our sampler can boost recognition performance in most settings.

In summary, our main contribution are as follows:
\begin{itemize}
    \item We propose a spatial-temporal sampler for few-shot action recognition with end-to-end differentiable implementations, which significantly improves video data utilization efficiency.
    \item We adopt a task-adaptive learner for our sampler, which can dynamically adjust the sampling strategy according to the episode task at hand.
    \item Our proposed sampler can be integrated with most current few-shot action recognition methods, demonstrating its versatility. Extensive experiments show a significant boost in their performances when leveraging our sampler.
\end{itemize}

%-------------------------------------------------------------------------
\section{Related Work}
%\textbf{Few-shot Learning} 
%Currently, metric-learning is the most popular solution to address the few-shot recognition task, which performs classification by comparing the seen and novel classes with learned metrics. Some notable works are discussed here. Matching network~\cite{matching2016} proposed an attention mechanism to calculate the similarities between samples based on the cosine metric. The RelationNet~\cite{relationNet2018} concatenates features of two samples and feeds them into the proposed relation network for comparison and classification instead of using specific metrics. The Prototypical Network~\cite{prototypical2017} aims to obtain a robust class representation by calculating a prototype for each category, and then it conducts nearest neighbor classification with these prototypes. Although the above methods have made significant progress in few-shot image classification, directly applying them to action recognition is less optimal.

\noindent
\textbf{Action Recognition and Frame Selection} 
Action recognition has made great progress with the development of 3D-CNN models. C3D~\cite{c3d2015} and I3D~\cite{I3d2017} extend traditional 2D CNN models to 3D versions to better conduct temporal modeling. Some approaches~\cite{p3d2017,r(2+1)d2018} decompose 3D-convolution into 2D \& 1D convolutions to learn spatial and temporal information separately, which yielded lower computational costs and better accuracy. Recently, some methods have attempted frame selection on input videos to improve inference efficiency. Since the selection operation is discrete, many methods utilize reinforcement learning~\cite{adaframe2018,adafocus2021,marl2019} or the Gumbel trick~\cite{arnet2020} to address this issue. Adaframe~\cite{adaframe2018} adaptively selects a small number of frames for each video for recognition, which is trained with the policy gradient method. MARL~\cite{marl2019} adopts multi-agent reinforcement learning to select multiple frames parallelly. FrameExit~\cite{frameexit2021} introduces the early stop strategy for frame sampling, and the frame selection stops when a pre-defined criterion is satisfied. Most of these works only focus on temporal selection. A recent work AdaFocus~\cite{adafocus2021} leverages reinforcement learning to locate a specific sub-region for each frame, and proceeds to pass it to classification. Both of these methods aim to reduce the number of frames to improve efficiency, but we are focused on selecting more representative frames at a fixed number to improve the utilization efficiency. Moreover, their methods are dependent on having sufficient training samples. Generally, we note that spatial sampling has not been well explored. More comparison and discussion between these methods will be detailed in Sec~\ref{sec:comparision}.

\noindent
\textbf{Few-shot Action Recognition} 
The majority of current studies on few-shot action recognition follow the metric learning paradigm. Due to the diverse distribution of actions, there may exist action misalignment among videos. Therefore, some alignment-based approaches are proposed to address this issue. TARN~\cite{tarn2019} proposed an attentive relation network to perform the temporal alignment implicitly at the video segment level. OTAM~\cite{otam2020} explicitly aligns video sequences with a variant of the Dynamic Time Warping (DTW) algorithm. TA$^2$N~\cite{ta2n2022} further deconstruct action misalignment into temporal and spatial aspects, and devises a joint spatial-temporal alignment framework to address them. Other methods focus on learning an enhanced feature representation for videos. An early work CMN~\cite{compound2018} proposed a compound memory network to store matrix representations of videos, which can be easily retrieved and updated. ARN~\cite{ARN2020} utilizes a self-supervised strategy to improve the robustness of video representation. TRX~\cite{trx2021} represents video by exhaustive pairs and triplets of frames to enlarge the metric space. However, all of them operate at the feature level while limited attention is paid to the importance of frame sampling.

\begin{figure*}[tp]
    \centering
    \includegraphics[width=1\linewidth]{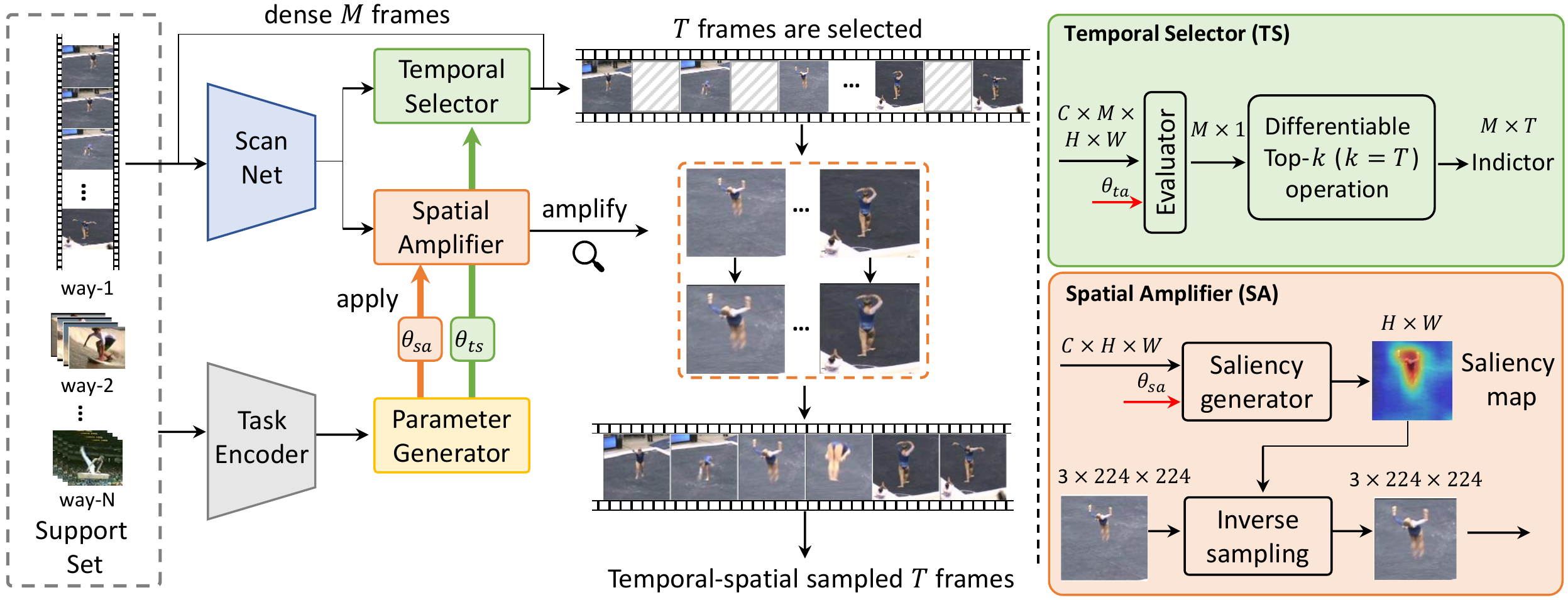}
    \caption{The framework of our proposed sampler. Videos are first sent to the Scan Network with dense frame sampling to get their global perception. Based on it, TS (temporal selector) evaluates frames and selects $T$ frames that contribute most to recognition. Then selected frames are processed by SA (spatial amplifier), which amplifies discriminative sub-regions under the guidance of saliency maps. An encoder aggregates the task information and generates task-specific parameters for TS and SA to conduct a dynamic sampling according to the episode task at hand.}
    \vspace{-0.5em}
    \label{fig:framework}
    \vspace{-0.5em}
\end{figure*}

\section{Methods}

\subsection{Problem definition} Following standard episode training in few-shot action recognition, the dataset is divided into three distinct parts: training set $\mathcal{C}_{train}$, validation set $\mathcal{C}_{val}$, and test set $\mathcal{C}_{test}$. The training set contains sufficient labeled data for each class while there exist only a few labeled samples in the test set. The validation set is only used for model evaluation during training. Moreover, there are no overlapping categories between these three sets. Generally, \emph{few-shot} action recognition aims to train a classification network that can generalize well to novel classes in the test set. Consider an $N$-way $K$-shot episode training, each episode contains a support set $\mathcal{S}$ sampled from the training set $\mathcal{C}_{train}$. It contains $N \times K$ samples from $N$ different classes where each class contains $K$ support samples. Likewise, $Q$ samples from each class are then selected to form the query
set $\mathcal{Q}$ which contains $N \times Q$ samples. The goal is to classify the $N \times Q$ query samples using only the $N \times K$ support samples. Fig.~\ref{fig:framework} presents the overall framework of our proposed sampler.

\subsection{Video scanning}
To get a global perception of video, we first \textbf{\textit{densely}} sample each video into $M$ frames. Thus, each video is represented by a dense frame sequence $\mathbf{X}=\{X_1, X_2, ..., X_M\}, |\mathbf{X}|=M$. Our sampler aims to conduct both temporal and spatial sampling among frames in $\mathbf{X}$ to obtain a subset $\mathbf{X}', |\mathbf{X}'|=T<M$. Given the dense sequence, a lightweight Scan Network $f(\cdot)$ takes its frames as input and obtains corresponding frame-level features $\displaystyle f_\mathbf{X} = \{f(X_1), f(X_2),\dots,f(X_M)\}\in \mathbb{R}^{C \times M\times H\times W}$. To reduce computation, each frame is down-scaled to a small size ($64\times$64 in default) before feeding into $f(\cdot)$. The video-level representation is determined as the average of all the frames $g_\mathbf{X} = \frac{1}{M}\sum_{i}^{M}f(X_i) \in \mathbb{R}^{C\times H \times W}$, for the subsequent sampling procedure. 

\subsection{Temporal selector}
A simple way to improve the data coverage and utilization efficiency is feeding all $M$ frames into few-shot learners without any filtering. However, this is intractable since it requires enormous computations and GPU memory, especially under episode training (each episode contains $N(Q+K)$ videos). Therefore, we aim to select $T$ informative frames that contribute the most to few-shot recognition from dense frames $\mathbf{X}$, which will be able to improve the data utilization without introducing further overhead. To this end, a temporal selector (TS) is devised to conduct this selection. 

\subsubsection{\textbf{Evaluation}}
Based on frame-level features, an evaluator $\Phi$ predicts the importance score for each frame. Specifically, it receives frame features and outputs importance scores $\mathbf{S}\in \mathbb{R}^{M\times 1}$. Meanwhile, the global information $g_\mathbf{X}$ is concatenated with each frame-level feature to provide a global perception:
\begin{gather}
    s_i = \Phi(\texttt{Avg}(\texttt{Cat}(f(X_i), g_\mathbf{X})) + \texttt{PE}(i)) \\
    \Phi = w_2(\texttt{ReLU}(w_1(\cdot))
\label{eq:evaluation}
\end{gather}
where the $w_1\in \mathbb{R}^{C\times d}$, $w_2 \in \mathbb{R}^{d\times 1}$ denotes the weights of linear layers, and $\texttt{Avg}$ denotes spatially global average pooling, $\texttt{PE}$ indicates the Position Embedding~\cite{transfomer2017} of position $i$. Specifically, the value of $w_2$ is dynamically adjusted among different episode tasks, which will be further elaborated in Sec.\ref{sec:task-ada}. Then, scores are then normalized to $[0,1]$ to stabilize the training process.

Based on the importance scores, we can pick the $T$ highest scores by a \texttt{Top-k} operation (set $k=T$), which returns the indices $\{i_1, i_2, \dots, i_T\}\in [0,M)$ of these $T$ frames. To keep the temporal order of selected frames, the indices are sorted s.t., $i_1<i_2<\dots<i_T$. Further, the indices are converted to one-hot vectors $\textbf{I} = \{I_{i_1}, I_{i_2}, \dots, I_{i_T}\}\in \{0,1\}^{M\times T}$. This way, the selected frame subset $X'$ could be extracted by matrix tensor multiplication:
\begin{gather}
    \mathbf{X}^{'} = \mathbf{I}^{T}\mathbf{X}
\end{gather}
Nevertheless, the above operations pose a great challenge that the \texttt{Top-k} and one-hot operations are \textbf{non-differentiable} for end-to-end training.

\subsubsection{\textbf{Differentiable selection}}
We adopt the \emph{perturbed maximum method} introduced in ~\cite{perturb2020,differential2021} to differentiate the sampling process during training. Specifically, the above temporal selection process is equivalent to solving a linear program of:
\begin{equation}
    \mathop{\arg\max}_{\mathbf{I}\in \mathcal{C}}\langle \mathbf{I}, \mathbf{S}\mathbf{1}^{T} \rangle
    \label{eq:linear}
\end{equation}
Here, input $\mathbf{S}\mathbf{1}^{T}\in \mathbb{R}^{M\times T}$ denotes repeating score $\mathbf{S}$ by $T$ times, while $\mathcal{C}$ indicates a convex constraint set containing all possible $\mathbf{I}$.
% \begin{equation}
% \begin{array}{r}
% \mathcal{C}=\left\{\mathbf{I} \in \mathbb{R}^{M \times T}: \mathbf{I}_{m, t} \geq 0, \mathbf{1}^{\mathbf{T}} \mathbf{I}=\mathbf{1}, \mathbf{I} \mathbf{1} \leq \mathbf{1}\right. \\
% \left.\sum_{i \in[M]} i \mathbf{I}_{i, t}<\sum_{j \in[M]} j \mathbf{I}_{j, t^{\prime}}, \forall t<t^{\prime}\right\}
% \end{array}
% \end{equation}
Under this equivalence, the linear program of Eq.\ref{eq:linear} can be solved by the perturbed maximum method, which performs forward and backward operations for differentiation as described below:

\noindent
\textbf{Forward} This step forwards a smoothed version of Eq.\ref{eq:linear} by calculating expectations with random perturbations on input:
\begin{equation}
    \mathbf{I}_{\sigma}=\mathbb{E}_{Z}\left[\underset{\mathbf{I} \in \mathcal{C}}{\arg \max }\left\langle\mathbf{I}, \mathbf{S 1}^{\mathbf{T}}+\sigma \mathbf{Z}\right\rangle\right]
\end{equation}
where $\sigma$ is a temperature parameter and $\mathbf{Z}$ is a random noise sampled from the uniform Gaussian distribution. In practice, we conduct the \texttt{Top-k} ($k=T$ in our case) algorithm with perturbed importance scores for $n$ (i.e. $n=500$ in our implementation) times, and compute their expectations.

\noindent
\textbf{Backward} Following \cite{perturb2020}, the Jacobian of the above forward pass can be calculated as:
\begin{equation}
    J_{\mathbf{s}} \mathbf{I}=\mathbb{E}_{Z}\left[\underset{\mathbf{I} \in \mathcal{C}}{\arg \max }\left\langle\mathbf{I}, \mathbf{S} \mathbf{1}^{\mathbf{T}}+\sigma \mathbf{Z}\right\rangle \mathbf{Z}^{\mathbf{T}} / \sigma\right]
\end{equation}
Based on this, we can back-propagate the gradients through the \texttt{Top-k} ($k=T$) operation.

During inference, we leverage hard \texttt{Top-k} to boost computational efficiency. However, applying hard \texttt{Top-k} during evaluation also creates inconsistencies between training and testing. To this end, we linearly decay $\sigma$ to zero during training. When $\sigma=0$, the differentiable \texttt{Top-k} is identical with the hard \texttt{Top-k}.

\subsection{Spatial amplifier}
Critical actions tend to occur in the partial regions of the frame, such as around the actors or objects. However, current approaches in few-shot action recognition treat these regions equal to other areas during data processing, thus reducing the efficiency of data utilization. Some works address this similar issue in image recognition by locating an important sub-region and then cropping it out from the whole image~\cite{crop1,racnn2017, scapnet2021}. However, directly replacing the original image with its partial region may further impact the data utilization efficiency under the few-shot setting. Inspired by application of attention-based non-uniform sampling~\cite{learningzoom2018,STN2015,trilinear2019}, we adopt a spatial amplifier (SA) for few-shot action recognition, which emphasizes the discriminative spatial regions while maintaining relatively complete frame information.

\begin{figure}[t]
    \begin{center}
        \includegraphics[width=1.0\linewidth]{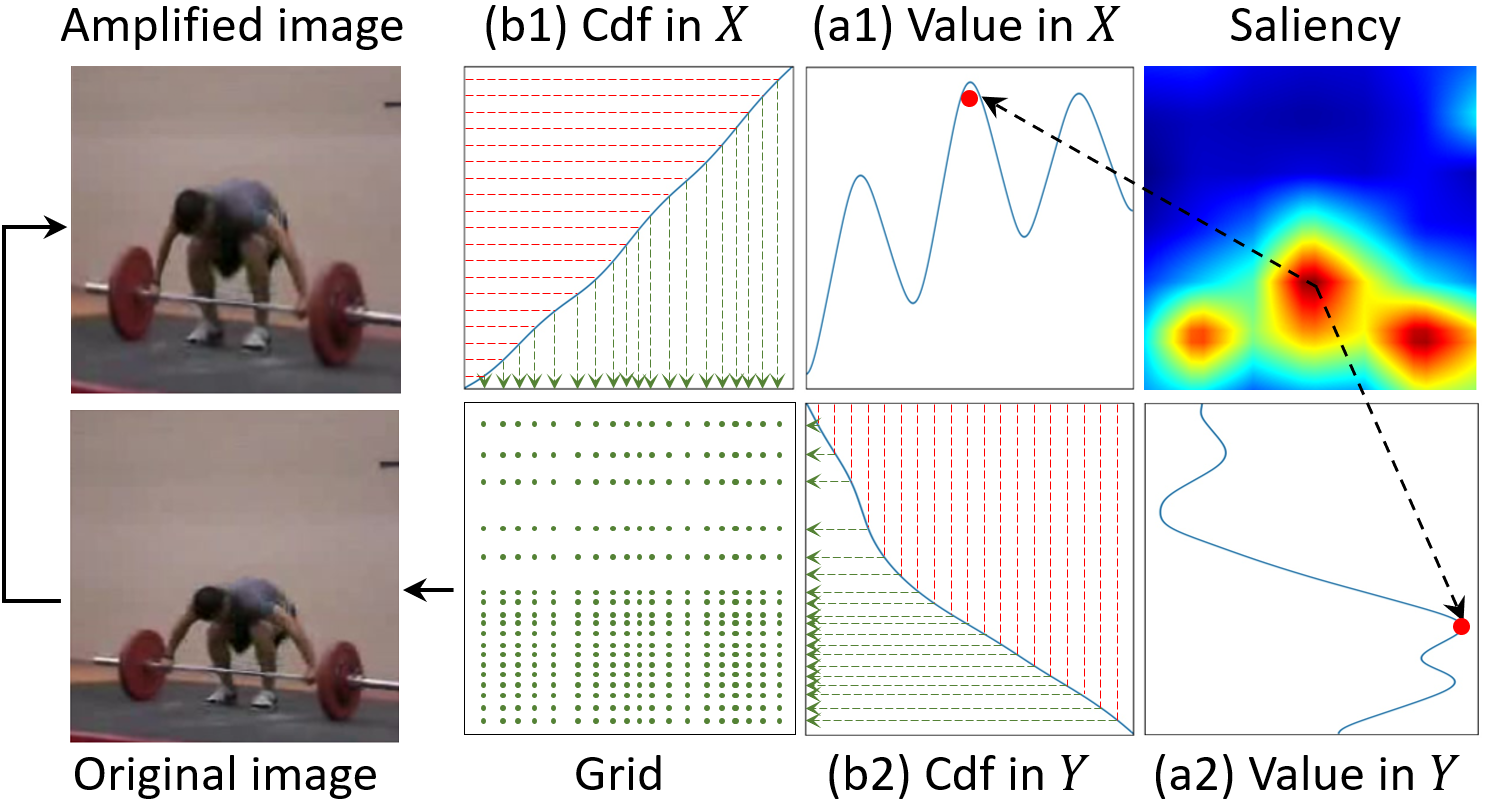}
    \end{center}
    \caption{Illustration of the 2D inverse-transform sampling.}
    %\vspace{-0.75em}
    \label{fig:spatial}
\end{figure}

\subsubsection{\textbf{Saliency map generation}} We first introduce how the informative regions of each frame are estimated. Since each feature map channel in CNN models may characterize a specific recognition pattern, we could estimate areas that contribute significantly to few-shot recognition by feature map aggregation. From the Scan Network $f(\cdot)$, we resort to using its output feature maps $f(X_i)\in \mathbb{R}^{C\times H \times W}$ to generate spatial saliency maps. However, there may exist activations in some action-irrelevant regions (e.g., background). Hence, directly aggregating all the channels will introduce some noises. Therefore, to enhance the discriminative patterns while dismissing these noises, we incorporate the self-attention mechanism~\cite{transfomer2017} on the channels of $f(X_i)$, such that
\begin{equation}
    \alpha = \frac{f(X)f(X)^T}{\sqrt{H\times W}} \in \mathbb{R} ^ {C\times C},~~f(X)'= \alpha f(X) \in \mathbb{R} ^ {C\times H\times W}
\end{equation}

Then, we can calculate a saliency map $\mathbf{M_s}$ for frame $i$ by aggregating the activations of all channels of the feature map:
\begin{equation}
    \mathbf{M_s}\in \mathbb{R}^{H\times W} = \frac{1}{C}\sum_{i}^{C} w_{s_i}\cdot f(X)'
\label{eq:saliency}
\end{equation}
where $w_s\in \mathbb{R}^{C\times 1}$ are learnable parameters that aggregate all the channels. Note that $w_s$ is also dynamically adjusted according to the episode task at hand. We elaborate more details in Sec.\ref{sec:task-ada}. Finally, saliency map $\mathbf{M}_s$ will be up-sampled to the same size of frame $X$.

\subsubsection{\textbf{Amplification by 2-D inverse sampling}} Based on the saliency map, our rule for spatial sampling is that an area with a large saliency value should be given a larger probability to be sampled (i.e., in this context, we say that this area will be `\emph{amplified}' compared to other regions). We implement the above amplification process by the \emph{inverse-transform sampling} used in ~\cite{inverse_sample1986, trilinear2019,saliencysampling2022}. 

As illustrated in \autoref{fig:spatial}, we first decompose the saliency map $\mathbf{M}_s$ into $x$ and $y$ dimensions by calculating the maximum values over axes following \cite{trilinear2019} for stabilization: 
\begin{gather}
    M_x = \max _{1 \leq i \leq W} (\mathbf{M_s})_{i, j},~~~~M_y = \max _{1 \leq j \leq H} (\mathbf{M_s})_{i, j}
\end{gather}
Then, we consider the Cumulative Distribution Function (\texttt{cdf}), which is non-uniform and monotonically-increasing, to obtain their respective distributions (Fig.\ref{fig:spatial} b1\&b2):
\begin{gather}
    D_x = \texttt{cdf}(M_x),~~~~D_y = \texttt{cdf}(M_y)
\label{eq:distribution}
\end{gather}
Therefore, the sampling function for frame $X$ under saliency map $\mathbf{M}_s$ can be calculated by the inverse function of Eq.\ref{eq:distribution}:
\begin{equation}
    X^{'}_{i,j} = \texttt{Func}(F, \mathbf{M_s}, i, j)=X_{{D}_{x}^{-1}(i), {D}_{y}^{-1}(j)}
\label{eq:sample_func}
\end{equation}
In practice, we implement Eq.\ref{eq:sample_func} by uniformly sampling points over the $y$ axis, and projecting the values to the $x$ axis to obtain sampling points (depicted by the green arrows shown in Fig.\ref{fig:spatial} b1\&b2). The sampling points obtained from Fig.\ref{fig:spatial} b1\&b2 form the 2D sampling grid. Finally, we can conduct an affine transformation above the original image based on the grid to obtain the final amplified image.

The SA performs on all frames selected by TS. In this way, the amplified frame can emphasize the discriminative regions while maintaining complete information of the original one.%\lhb{as Fig.xx} %global information of each frame is relatively complete. 

\begin{figure}[t]
    \begin{center}
        \includegraphics[width=1\linewidth]{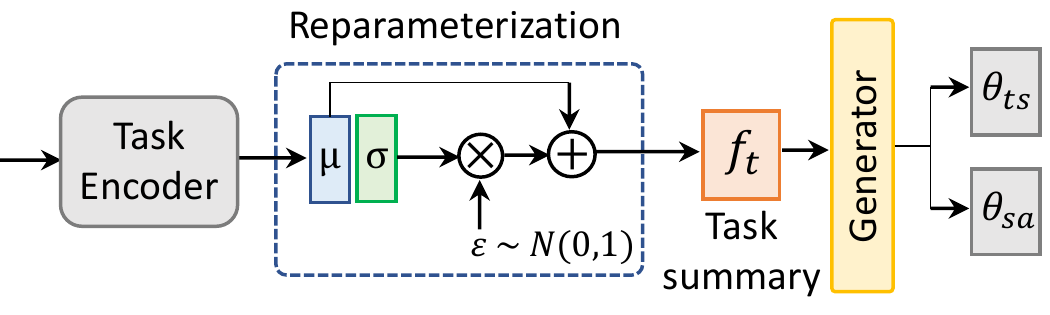}
    \end{center}
    \caption{Illustration of the task-adaptive sampling learner.}
    %\vspace{-0.75em}
    \label{fig:task}
\end{figure}
\subsection{Task-adaptive sampling learner}
\label{sec:task-ada}
In general action recognition, once a sampler is well-trained, it samples each testing video with a fixed strategy and criterion~\cite{adaframe2018,arnet2020}. Nevertheless, in the few-shot episode paradigm, query video relies on videos in support set at hand to conduct classification. Thus, our testing videos are not independent compared to general action recognition. Therefore, fixing the sampling strategy for each video among episodes is not ideal in few-shot recognition. To this end, we adopt a task-adaptive learner for our sampler, which generates task-specific parameters for layers in TS and SA to dynamically adjust the sampling strategy according to the episode task at hand. %Specifically, it can be implemented by generating dynamic parameters for TS and SA.

Given a support task $\mathcal{S}=\{(\mathbf{X}^i,\mathbf{Y}^i)_{i}^{N\times K}\}$ and its corresponding video-level features $\{(g_{\mathbf{X}^i})_{i}^{N\times K}\}$  (extracted by Scan Net) at hand, we can estimate the summary statistics of this task by parameterizing it as a conditional multivariate Gaussian distribution with a diagonal covariance~\cite{static2016,LGMNet2019}. Therefore, a task encoder $E$ is employed to estimate its summary statistics:
\begin{equation}
    \mu, \sigma=\frac{1}{N\times K}\sum_{i=1}^{N\times K}E(\texttt{Avg}(g_{\mathbf{X}^i}))
\end{equation}
where $E$ consists of two linear layers, $\texttt{Avg}$ is global average pooling,  $\mu$ and $\sigma$ are 128-$dim$ mean and variance estimated by $E$. Then, we denote the probability distribution of task summary feature as:
\begin{equation}
\small
    p(f_t \mid \mathcal{S})=\mathcal{N}\left(\mu, \operatorname{diag}\left(\sigma^{2}\right)\right)
\end{equation}
We could sample a task summary feature $f_{t}$ that satisfies the above distribution with the \textit{re-parameterization} trick~\cite{vae2013}: $f_{t} \in \mathbb{R}^{128} = \mu + \sigma \varepsilon$, where $\varepsilon$ is a random variable that $\varepsilon\sim\mathcal{N}(0,1)$. In this way, each task is encoded into a fixed-length representation, which reflects more consistently across the 
%its unique 
data distribution of task.

Based on the summary of current task, a task-specific sampling strategy can be implemented by adjusting the parameters that influence the criterion of sampling. Specifically, we generate task-specific parameters for $w_{2}\in \mathbb{R}^{d\times 1}$ in TS (Eq.\ref{eq:evaluation}) and $w_s\in\mathbb{R}^{C\times 1}$ in SA (Eq.\ref{eq:saliency}) where the former decides how to evaluate the importance score of frames while the latter determines the generation of saliency maps. The above process is illustrated in Fig.\ref{fig:task}.
\begin{equation}
    \theta_{ts} = G_t(f_t)  ,~~\theta_{sa} = G_s(f_t) 
\end{equation}
where $G_t\in \mathbb{R}^{128\times d}$ and $G_s\in \mathbb{R}^{128\times C}$ denotes the weights of the parameter generator. Finally, the generated parameters are normalized and filled into corresponding layers ($w_2 \leftarrow \frac{\theta_{ts}}{\Vert \theta_{ts} \Vert}$, $w_s\leftarrow \frac{\theta_{sa}}{\Vert \theta_{sa} \Vert}$).

\subsection{Optimization}
Since our sampler can be integrated with most existing few-shot action recognition methods, we use the strong baseline ProtoNet~\cite{prototypical2017} in default for the optimization. Our sampler is trained with the baseline network in an end-to-end manner. All query and support videos (denoted as $\mathbf{X}_q$ and $\mathbf{X}_s$) are first pre-processed by our sampler to get sampled input $\mathbf{X}_q^{'}$ and $\mathbf{X}_s^{'}$. Then, they are passed through a ResNet-50~\cite{he2016deep} for feature extraction and prototype learning. Given the feature $F_q$ of query video, and the support prototype $F_s^c$ 
of class $c$, we concatenate each of them with the corresponding features $f_q$ and $f_s^c$ (global feature obtained by scan-Net $f(\cdot)$ during sampling) to form the final feature for classification:
\begin{equation}
    F_q = w_{r}(\texttt{Cat}(F_q, f_q)),~F_s^c = w_{r}(\texttt{Cat}(F_s^c, f_s^c))
\end{equation}
where linear layer $w_{r}$ reduces their feature dimension into 2048-dim.
Therefore, the classification probability is:
\begin{equation}
    {\rm P}(\mathbf{X}_q \in c_i)=\frac{\exp(-dist(F_q,F_s^{c_i}))}{\sum_{c_j \in C}{\exp(-dist(F_q,F_s^{c_j}))}}
\end{equation}

\begin{equation}
    dist(F_q,F_s^c)=\sum_{t=1}^{T}{1- \frac{\langle F_{q,[t]},F^c_{s,[t]}\rangle}{\|F_{q,[t]}\|_2 \|F^c_{s,[t]}\|_2} },
\end{equation}
where $dist(f,p)$ is the frame-wise cosine distance metric.
Then. the classification loss is calculated with cross-entropy:
\begin{equation}
    \mathcal{L}_{CE}= -\sum_{q\in Q}\mathbf{Y}_q\cdot\log{{\rm P}(\mathbf{X}_q \in c_i)},
\end{equation}
where $\mathbf{Y}_q \in \{0,1\}$ indicates whether $\mathbf{X}_q\in c_i$, while $Q$ and $C$ represent the query set and its corresponding class label set.

Besides, to make sure that our sampler can well capture the informative frames for few-shot recognition, we add a classification loss on our sampler to provide intermediate supervision. Specifically, an auxiliary classification loss $\mathcal{L}_{aux}$ is calculated using the features $f_q, f_s$ (output by the scan-Net $f(\cdot)$) in the same way as $\mathcal{L}_{CE}$. Note that only the features of selected $T$ frames are involved in this loss calculation. Hence, the overall loss for optimization is:
\begin{equation}
    \mathcal{L} = \mathcal{L}_{CE} + \mathcal{L}_{aux}
\end{equation}

\begin{table*}[t]
\centering
\small
%\scriptsize
% \setlength{\tabcolsep}{0.5em}
\begin{tabular}{c|c|cc|cc|cc|cc}
\hline
\multirow{2}{*}{Method} &
\multirow{2}{*}{Backbone} &
\multicolumn{2}{c|}{HMDB51} & \multicolumn{2}{c|}{UCF101} & \multicolumn{2}{c|}{SSv2} & \multicolumn{2}{c}{Kinetics-CMN} \\
 & & 1-shot & 5-shot & 1-shot & 5-shot & 1-shot & 5-shot & 1-shot & 5-shot \\ \hline
CMN~\cite{cmn-j2020} & ResNet-50 &- & - & - & - & - & - & 60.5 & 78.9 \\
TARN~\cite{tarn2019} & C3D& - & - & - & - & - & - & 64.8 & 78.5 \\
ARN~\cite{ARN2020} & 3D-\textit{464}-Conv& 45.5 & 60.6 & 66.3 & 83.1 & - & - & 63.7 & 82.4 \\
TRN++~\cite{otam2020} & ResNet-50& - & - & - & - & 38.6 & 48.9 & 68.4 & 82.0 \\
ProtoNet*~\cite{prototypical2017} & ResNet-50& 54.2 & 68.4 & 78.7 & 89.6 & 39.3 & 52.0 & 64.5 & 77.9 \\
OTAM~\cite{otam2020} & ResNet-50& 54.5 & 66.1 & 79.9 & 88.9 & 42.8 & 52.3 & 73.0 & 85.8 \\
TRX~\cite{trx2021} & ResNet-50& 53.1 & 75.6 & 78.6 & 96.1 & 42.0 & \textbf{64.6} & 63.6 & 85.9 \\
TA$^2$N~\cite{ta2n2022} & ResNet-50& 59.7 & 73.9 & 81.9 & 95.1 & \textbf{47.6} & 61.0 & 72.8 & 85.8 \\
\hline
ProtoNet + Sampler & ResNet-50& 59.0 (\up{4.8$\uparrow$}) & 72.8 (\up{4.4$\uparrow$}) & 82.5 (\up{3.8$\uparrow$}) & 93.6 (\up{4.0$\uparrow$}) & 41.8 (\up{2.5$\uparrow$}) & 57.0 (\up{5.0$\uparrow$}) & 71.0 (\up{6.5$\uparrow$}) & 84.8 (\up{6.9$\uparrow$}) \\
OTAM + Sampler & ResNet-50& 58.0 (\up{3.5$\uparrow$}) & 72.2 (\up{6.1$\uparrow$}) & 81.6 (\up{1.7$\uparrow$}) & 93.7 (\up{4.8$\uparrow$}) & 43.0 (\up{0.2$\uparrow$}) & 56.4 (\up{4.1$\uparrow$}) & \textbf{74.5} (\up{1.5$\uparrow$})& 85.3 (\down{0.5$\downarrow$}) \\
TRX + Sampler & ResNet-50&  53.5 (\up{0.4$\uparrow$})   &   \textbf{76.0} (\up{0.4$\uparrow$}) &   82.1 (\up{3.5$\uparrow$}) &  \textbf{96.1} (\up{0.0$\uparrow$})  &  43.6 (\up{1.6$\uparrow$})   &  64.4 (\down{0.2$\downarrow$})   &  65.8 (\up{2.2$\uparrow$})   &  86.0 (\up{0.1$\uparrow$})  \\
TA$^2$N + Sampler & ResNet-50& \textbf{59.9} (\up{0.2$\uparrow$}) & 73.5 (\down{0.4$\downarrow$})   & \textbf{83.5} (\up{1.6$\uparrow$})& 96.0 (\up{0.9$\uparrow$}) & 47.1 (\down{0.5$\downarrow$})   &  61.6 (\up{0.6$\uparrow$})   &   73.6 (\up{0.8$\uparrow$})  &  \textbf{86.2} (\up{0.4$\uparrow$})   \\ \hline
\end{tabular}
\caption{Few-shot action recognition results under standard 5-way k-shot setting. * refers to our re-implementation.}
\label{tab:compare}
%\vspace{-0.75em}
\end{table*}
\section{Experiments}
\subsection{Datasets and baselines}

\noindent
\textbf{Datasets} Four popular datasets are selected for experiments:
\begin{itemize}
    \item \textit{UCF101}~\cite{ucf101}:
    We follow the same protocol introduced in ARN~\cite{ARN2020}, where 70/10/21 classes and 9154/1421/2745 videos are included for train/val/test respectively.
    \item \textit{HMDB51}~\cite{HMDB}:  We also follow the protocol of ARN~\cite{ARN2020}, which takes 31/10/10 action classes with 4280/1194/1292 videos for train/val/test.
    \item \textit{SSv2}~\cite{ssv2}: We adopt the same protocol as OTAM~\cite{otam2020} where 64/12/24 classes and 77500/1925/ 2854 videos are included for train/val/test respectively.
    \item \textit{Kinetics-CMN}~\cite{compound2018} contains 100 classes selected from Kinetics-400, where 64/12/24 classes are split into train/val/test set with 100 videos for each class.
\end{itemize}

\noindent
\textbf{Baselines} The strong and widely-used FSL baseline ProtoNet~\cite{prototypical2017} is selected as our main baseline. Besides, we choose recent state-of-the-art FSL action recognition approaches, including CMN-J~\cite{cmn-j2020}, ARN~\cite{ARN2020}, TARN~\cite{tarn2019}, OTAM~\cite{otam2020}, TRX~\cite{trx2021}, and TA$^{2}$N~\cite{ta2n2022}. Among them, we integrate our sampler with ProtoNet, OTAM, TRX, and TA$^{2}$N to validate our effectiveness.

\subsection{Implementation details}
We use the ShuffleNet-v2~\cite{shufflenet2018} pre-trained in ImageNet as the Scan Network for efficiency. In addition, we remove its first max-pooling layer to improve the resolution of saliency maps. By default, we scan each video by 
%with dense frame sequence, by default, we 
densely sampling 16 frames at $224\times 224$ pixels (i.e., $M=16$). % for each video.
Moreover, each frame is downscaled to $64\times 64$ pixels before feeding into the Scan Network. Following previous works, our sampler then outputs 8 frames (i.e., $T=8$) at $224\times 224$ pixels for the subsequent few-shot learners to ensure fair comparisons. For episode training, 5-way 1-shot and 5-way 5-shot classification tasks are conducted. The whole meta-training ran for 15,000 epochs, and each epoch consists of 200 episodes. In the testing phase, we sample 10,000 episodes in the meta-test split and report the average result. Training is optimized by SGD with momentum, and the initial learning rate is 0.01, which is decayed (multiplied) by 0.9 after every 5000 epochs. The temperature $\sigma$ is set to 0.1, which is decayed by 0.8 after 2000 epochs. When we plug our sampler into other methods (e.g., OTAM, TRX), we follow their default settings and models, conducting end-to-end training with our sampler. 

\begin{figure*}[t]
    \centering
    \includegraphics[width=1.0\linewidth]{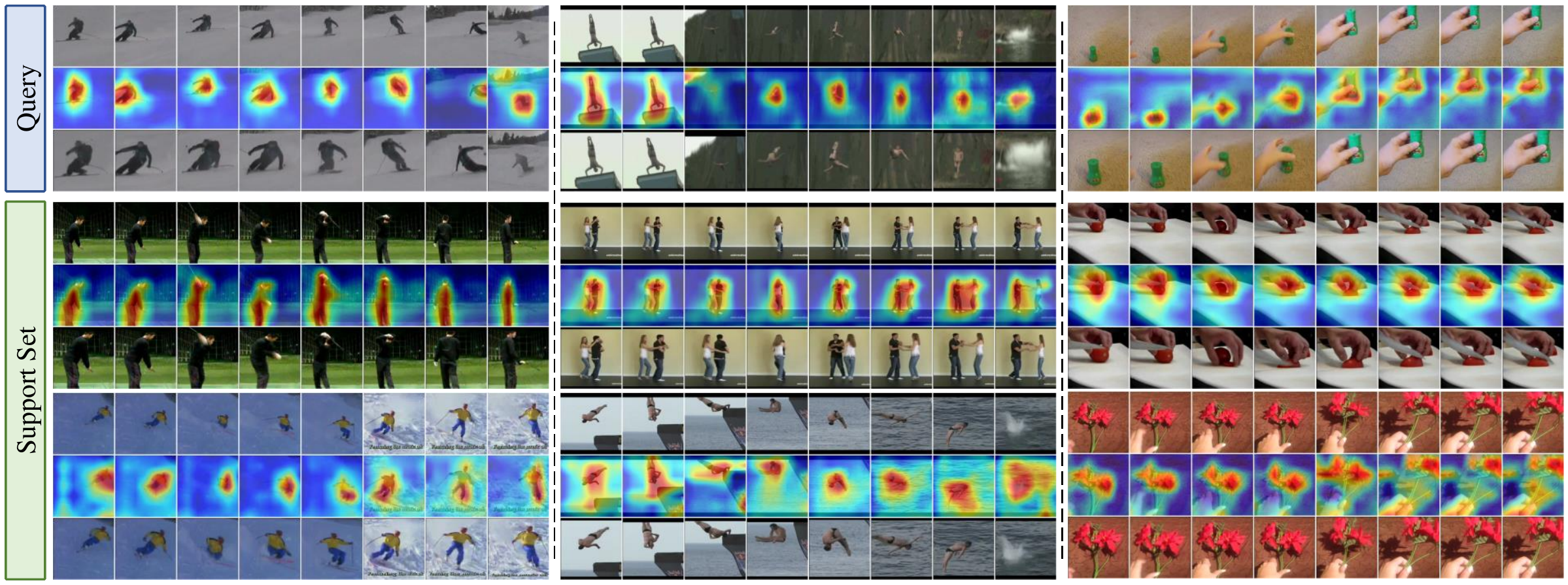}
    \vspace{-2em}
    \caption{Visualization of results of the proposed sampler in three sample episode tasks from UCF and SSv2 datasets. The \textit{1st} row: frames selected by TS, the \textit{2nd} row: saliency maps, the \textit{3rd} row: frames amplified by SA. 2-way 1-shot is illustrated here.
    }
    \label{fig:vis}
    %\vspace{-1em}
\end{figure*}

\subsection{Main results}
\subsubsection{\textbf{Quantitative results}}
\autoref{tab:compare} summarizes the results of the above methods on few-shot action recognition. It is clear that our sampler consistently boosts the performance of various baselines in most cases. From appearance-dominated video benchmarks (HMDB, UCF, and Kinetics) to temporal-sensitive datasets (SSv2), we demonstrate the benefits of our proposed sampler. When our sampler is adopted in the simple ProtoNet baseline, its performance is significantly improved to a level that is competitive with current state-of-the-art methods. For alignment-based methods (OTAM, TA$^2$N), we also observe promising improvements to all datasets brought upon by our sampler. While TRX is currently regarded as the best method in the 5-shot setting (but with a less ideal 1-shot performance), our sampler still yields a great improvement over the TRX baseline on all cases, especially in the 1-shot setting. We note that the observed improvement can be directly attributed to the application of our sampler, and not other training tricks or factors as all conditions are held fixed. These results demonstrate the effectiveness and generalization of our proposed sampler. 
%Besides, we also notice that the performance slightly drops over OTAM in 5-shot setting on Kinetics.
\subsubsection{\textbf{Qualitative visualization}}
Fig.~\ref{fig:vis} visualizes the temporal-spatial sampling results for videos in episode tasks. We observe that TS can select frames containing crucial and complete action processes (e.g., selected frames for `cliff diving' in col.2). Moreover, our generated salience map complements by indicating the important spatial regions for each frame. Accordingly, as the results of SA show, these regions with higher saliency are well amplified and emphasized (e.g., the person in col.1st and the hand in col.3rd), which in turn, makes them easier to be recognized. Meanwhile, the amplified images still maintain relatively complete information 
of the original ones. In summary, the visualization tellingly depicts the effectiveness of our sampling strategy in temporal and spatial dimensions. More visualizations are presented in our supplementary.
\begin{table}[t]
\small
\centering
\begin{tabular}{c|cc}
\hline
\multirow{2}{*}{Method} & \multicolumn{2}{c}{ActivityNet} \\ \cline{2-3} 
                         & 1-shot         & 5-shot         \\ \hline
ProtoNet*                 &      66.5          &       80.3         \\
OTAM*                    &       69.9        &         85.2       \\
TRX*                     &       69.9        &        88.7        \\ \hline
ProtoNet + Sampler          &      73.5 (\up{7.0$\uparrow$})          &         86.8 (\up{6.5$\uparrow$})      \\
OTAM + Sampler              &        \textbf{74.7} (\up{4.8$\uparrow$})       &       87.5 (\up{2.3$\uparrow$})        \\
TRX + Sampler               &           72.0 (\up{2.1$\uparrow$})    &        \textbf{89.9} (\up{1.2$\uparrow$})        \\ \hline
\end{tabular}
\caption{Results on ActivityNet. * means results from our re-implementation.}
\label{tab:activity}
\vspace{-3em}
\end{table}
\subsubsection{\textbf{Effectiveness on long-term video}} Current few-shot action recognition methods only conduct evaluation on short-term videos (e.g., UCF, HMDB, Kinetics). In this paper, we benchmark various methods with our sampler on the long-term video dataset: ActivityNet~\cite{activity2015}, which is the first attempt to our knowledge. Following OTAM~\cite{otam2020}, we randomly selected 100 classes from the whole dataset. The 100 classes are then split into 64, 12, and 24 classes as the meta-training, meta-validation, and meta-testing sets, respectively. 70 samples are randomly selected for each class. ~\autoref{tab:activity} summarizes the experimental results on ActivityNet. We observe that the existing SOTA methods do not achieve satisfying results on such long-term videos when compared against the simple baseline ProtoNet, especially on the 1-shot setting. The reason may be that the ignoring of critical action caused by uniform sampling is more severe in such long-term videos. Moreover, each video in ActivityNet may contain multiple action segments, further increasing the difficulty of metric learning. When equipping these methods with our sampler, their performance can be boosted by a significant margin. This demonstrates that conducting sampling on such long-term and multi-activity videos is necessary and beneficial.

\begin{table}[t]
\small
\centering
\begin{tabular}{ccc|cc|cc}
\hline
\multirow{2}{*}{TS}  & \multirow{2}{*}{SA}  & \multirow{2}{*}{ada}  & \multicolumn{2}{c|}{UCF101}                     & \multicolumn{2}{c}{SSv2}                    \\
                     &                      &                       & 1-shot               & 5-shot                & 1-shot               & 5-shot               \\ \hline
\multicolumn{1}{l}{} & \multicolumn{1}{l}{} & \multicolumn{1}{l|}{} & 78.7 & 89.6 & 39.3 & 52.0 \\
\checkmark                    &                      &                       & 80.2                 & 91.4                  & 40.9                 & 55.6                 \\
                     & \checkmark                    &                       & 80.7                 & 92.1                  & 39.5                 & 53.5                 \\
\checkmark                    & \checkmark                    &                       & 81.8                 & 92.9                  & 41.5                 & 55.9                 \\
\checkmark                    &                      & \checkmark                     & 81.1                 & 92.2                  & 41.3                 & 56.4                 \\
                     & \checkmark                    & \checkmark                     & 81.6                 & 92.7                  & 39.8                 & 53.7                 \\
\checkmark                    & \checkmark                    & \checkmark                     & \textbf{82.5}                 & \textbf{93.6}                 & \textbf{41.8}                 & \textbf{57.0}                 \\ \hline
\end{tabular}
\caption{Ablation of each core component. TS: temporal selector; SA: spatial amplifier; ada: task-adaptive learner.}
\label{tab:breakdown}
\vspace{-2.5em}
\end{table}

\begin{table}[t]
\small
\centering
\begin{tabular}{c|cccccc}
\hline
M =      & 12   & 16   & 20   & 24   & 28   & 32   \\ \hline
UCF101 & 81.8 & \textbf{82.5} & 82.5 & 82.0 & 81.8 & 81.9 \\ \hline
SSv2   & 40.7 & 41.8 & \textbf{42.2} & 41.8 & 41.0 & 40.9 \\ \hline
\end{tabular}
\caption{Ablation of frequency of dense sampling, reported under 5-way 1-shot setting.}
\label{tab:frequency}
\vspace{-3.5em}
\end{table}
\subsection{In-depth study}
The following experiments are conducted based on the `ProtoNet + Sampler' model in default.
\subsubsection{\textbf{Breakdown analysis}} 
In \autoref{tab:breakdown}, we first conduct an ablation study to illustrate the effect of each core component in our sampler. All modules yield a stable improvement over the baseline. The TS brings significant gains on UCF and SSv2, indicating that conducting temporal selection is crucial for few-shot action recognition. Besides, we also observe that the SA works better on UCF than SSv2 dataset. Since the videos of SSv2 are first-person scenes and the main object in frames are human hands (as shown in 3rd col in Fig.\ref{fig:vis}), amplifying such hands leads to limited improvement in recognition. Further, by introducing task adaptation for TS and SA, they were able to reach their peak performances. It demonstrates that adjusting the sampling strategies among episode tasks is indeed beneficial and essential for few-shot action recognition.

\begin{table}[t]
\small
\centering
\begin{tabular}{c|cccccc}
\hline
T =     & 4    & 6    & 8    & 10   & 12   & 14   \\ \hline
UCF101 & 80.9 & 81.4 & 82.5 & 82.8 & \textbf{83.3} & 82.6 \\ \hline
SSv2   & 38.6 & 40.3 & 41.8 & 42.5 & 43.7 & \textbf{44.4} \\ \hline
\end{tabular}
\caption{Ablation of number of sampled frames, reported under 5-way 1-shot setting.}
\label{tab:number_of_T}
\vspace{-3em}
\end{table}
\subsubsection{\textbf{Frequency of scanning}}
To make sure that our Scan Network can cover more frames in the videos, we set $M=16$ by default to provide a trade-off between computation and performance. To investigate the effect of video scanning frequency in our sampler, we adjust different scanning frequencies (i.e., value of M) for experiments. Results are shown in \autoref{tab:frequency}. Surprisingly, directly increasing the frequency cannot guarantee a consistent improvement to the baselines. When enlarging $M$ to 20, the performance of SSv2 further improved, while there were no gains on the UCF dataset. However, when $M$ reaches $24\&28$, their improvements had stagnated %cannot be further improved. 
while some of them even encountered slight drops. There exists a large decrease in the performance of UCF and SSv2 when $M=32$. The reason may be that learning a good selection from $32$ elements is difficult using only a few training samples.

\subsubsection{\textbf{How many frames do we need?}}
For fair comparisons, we follow existing methods to sample $T=8$ frames as input for few-shot learners in \autoref{tab:compare}. However, by virtue of our sampler, increasing the number of sampling is likely to introduce more informative frames for recognition instead of redundancy. We adjust the $T$ from $6$ to $14$ to explore this issue in \autoref{tab:number_of_T}. For UCF101, the performance grows with $T$, and it peaks at $T=12$. However, its performance slightly drops when $T$ increases to 14. For SSv2, it is clear that increasing the number of sampled frames can consistently boost the performance. On the contrary, when reducing the frame number to $4$, the SSv2 suffers a significant drop in performance while UCF101 only drops slightly. This is consistent with the observation that SSv2 is more sensitive to temporal information.

\subsubsection{\textbf{t-SNE visualization}} 
\begin{figure}[t]
    \begin{center}
        \includegraphics[width=0.9\linewidth]{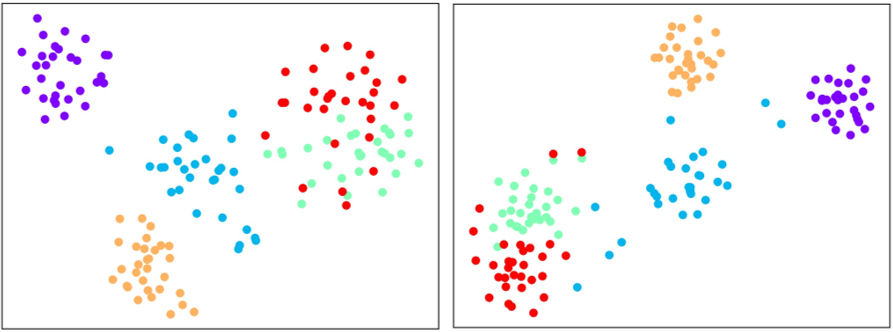}
    \end{center}
    \vspace{-1em}
    \caption{t-SNE feature embedding of videos w/o (left) and with (right) our proposed sampler. Reported using all videos in a 5-way 30-shot episode on ActivityNet.}
    \vspace{-0.5em}
    \label{fig:tsne}
\end{figure}

To illustrate the effect of our proposed sampler at the feature level, we visualize the feature embedding of videos with and without our sampler using t-SNE in \autoref{fig:tsne}. We can see that each cluster appears more concentrated after sampling. It further proves that our sampler can facilitate better feature metric learning.

%\subsubsection{\textbf{Distribution of sampled frames}}
%To gain more intuitive insights, we keep count of the temporal locations of the selected frames by TS in~\autoref{fig:distribution}. We observe that the distribution varies in these two datasets. For UCF101, our sampler tends to select frames from the beginning, middle, and end of a video, especially the middle part. This is reasonable because frames in the middle of sports videos are more likely to contain key actions. For SSv2, it tends to observe more frames at the beginning of videos. Besides, it will select some frames in the middle and end for final decisions. We reason that the classifier needed to observe the direction of action at the start of the video to distinguish fine-grained action categories in SSv2 (e.g., `\textit{Moving something up}' and `\textit{Moving something down}').
% \begin{figure}[t]
%     \begin{center}
%         \includegraphics[width=0.9\linewidth]{./fig/distribution.pdf}
%     \end{center}
%     \vspace{-1em}
%     \caption{Histogram of sampled frame locations on the meta-test set of UCF101 and SSv2. Reported with $T=8$ and $M=16$.}
%     \vspace{-1em}
%     \label{fig:distribution}
% \end{figure}

\subsubsection{\textbf{Solutions on spatial sampling}} In the image classification, there exists another popular solution for spatial sampling, which is to locate a sub-region and then crop it from the original image~\cite{crop1,racnn2017}. However, this operation may not be ideal for few-shot learning since it reduces the data utilization efficiency in the spatial dimension. Moreover, it requires setting a fixed size for the sub-region to achieve differentiable implementation and often involves multi-stage training. In contrast, our spatial amplifier can emphasize the discriminative region while maintaining most information of complete images. Besides, we can zoom the sub-regions with a flexible ratio. To gain further insight, we replace our SA with a crop-based solution (refer to supplementary for details). Results are summarized in \autoref{tab:crop}. We see that the crop-based solution only improves the performance of SSv2 by a slight margin. Moreover, the performance of UCF drops using crop-based spatial sampling while our amplifier brings a promising improvement on all baselines.

\subsubsection{\textbf{Analysis of task-adaptive learner}} As presented in \autoref{tab:breakdown}, the advantages of TS and SA are further enhanced by the task-adaptive sampling strategy, which has proven that dynamic sampling is beneficial to few-shot learning. Besides, we notice that it brings a greater improvement on the 5-shot setting than the 1-shot on all datasets. A possible reason is that the encoder can estimate more precise summary statistics from the 5-shot than the 1-shot as the number of samples increases. More analyses about our task-adaptive learner are provided in the supplementary.
\begin{table}[t]
\small
\centering
\begin{tabular}{c|cc|cc}
\hline
\multirow{2}{*}{Style} & \multicolumn{2}{c|}{UCF101} & \multicolumn{2}{c}{SSv2} \\
                       & 1-shot      & 5-shot     & 1-shot      & 5-shot     \\ \hline
None                & 81.8        & 92.2       & 41.3        & 56.4      \\
Amplifier              & \textbf{82.5 }       & \textbf{93.6}       & \textbf{41.8}        & \textbf{57.0}       \\
Crop                   & 81.3        & 92.0       & 41.4        & 56.6       \\ \hline
\end{tabular}
\caption{Results of different solution for spatial sampling. `None' means no spatial operation is performed.}
\label{tab:crop}
\vspace{-2em}
\end{table}

\begin{table}[t]
\centering
\small
\begin{tabular}{c|c|c}
\hline
Model            & Params & Speed      \\ \hline
ProtoNet         & 25.8 M & 0.84 s/task \\ \hline
ProtoNet + Sampler & 27.3 M \down{(1.5M$\uparrow$)} & 0.98 s/task \down{(0.14s$\uparrow$)} \\ \hline
\end{tabular}
\caption{Overhead analysis. Reported under 5-w 1-s task.}
\label{tab:overhead}
\vspace{-3em}
\end{table}
\subsubsection{\textbf{Comparison with sampler in action recognition}}
\label{sec:comparision}
As we stated in the related work, there exist some works~\cite{marl2019,arnet2020,frameexit2021} on general action recognition that conduct frame selection. However, the main difference is that they aim to sample fewer frames mainly to improve inference speed while we seek to sample \emph{more informative} frames under a fixed number of sampled frames. Moreover, most of them focus on temporal selection, and very few explorations were dedicated to the spatial dimension. Further, most of them involve reinforcement learning~\cite{adafocus2021,adaframe2018,marl2019} or Gumbel trick~\cite{arnet2020} for differential training. The former is intractable to train, while the latter cannot address the fixed-size subset selection issue during training. Besides, these methods also require large training samples. Most importantly, their sampling strategies are fixed during inference, while we can dynamically adjust sampling strategies according to the episode task at hand. For an extensive analysis of these differences, we also apply some of them to few-shot action recognition and compare their performances against ours (please refer to the supplementary for more details).

\subsubsection{\textbf{Overhead analysis}} As shown in \autoref{tab:overhead}, our sampler only adds less than 5\% parameter overhead, and the inference speed is still fast compared to basic few-shot learners. This demonstrates that our sampler is low-cost and lightweight.

\section{Conclusion}
We propose a novel task-adaptive spatial-temporal video sampler for few-shot action recognition. The sampler contains a temporal selector and a spatial amplifier which works hand-in-hand. The temporal selector conducts informative frame selection from the whole video with a differentiable implementation. The spatial amplifier emphasizes the discriminative features by amplifying the critical sub-regions in frames. Moreover, the sampling strategy is also dynamically adjusted according to the episode task at hand. Extensive experiments affirm that our proposed sampler can improve the performance of current methods by a significant margin.

\section{ACKNOWLEDGMENTS}
This paper is supported in part by the following grants: National Key Research and Development Program of China Grant (No.2018AAA0\\100400), National Natural Science Foundation of China (No.U21B2013, 61971277).

\balance
\bibliographystyle{ACM-Reference-Format}
\bibliography{egbib}

\clearpage
\appendix
\section{More ablation study}
% \subsection{Distribution of sampled frames}
% To gain more intuitive insights, we keep count of the temporal locations of the selected frames by TS in~\autoref{fig:distribution}. We observe that the distribution varies in these two datasets. For UCF101, our sampler tends to select frames from the beginning, middle, and end of a video, especially the middle part. This is reasonable because frames in the middle of sports videos are more likely to contain key actions. For SSv2, it tends to observe more frames at the beginning of videos. Besides, it will select some frames in the middle and end for final decisions. We reason that the classifier needed to observe the direction of action at the start of the video to distinguish fine-grained action categories in SSv2 (e.g., `\textit{Moving something up}' and `\textit{Moving something down}').
% \begin{figure}[t]
%     \begin{center}
%         \includegraphics[width=0.9\linewidth]{./fig/distribution.pdf}
%     \end{center}
%     \vspace{-1em}
%     \caption{Histogram of sampled frame locations on the meta-test set of UCF101 and SSv2. Reported with $T=8$ and $M=16$.}
%     \vspace{-1em}
%     \label{fig:distribution}
% \end{figure}

\subsection{Frame selector in general action recognition} As we have discussed in our paper, most existing frame selection methods cannot directly apply to address our issues (fix-size subset selection). To this end, we apply the recently proposed MGSampler~\cite{mgsampler} to few-shot action recognition, which can be regarded as an independent pre-processing operation for video frame sampling. Specifically, it first calculates the feature difference between adjacent frames to estimate the motion process for the whole video. Then, frame selection is performed based on this estimated motion information. In this way, we feed the $T=8$ frames selected by MGSampler based on the calculated motion difference among $M=20$ frames into ProtoNet to report its performance. As shown in \autoref{tab:general}, the MGSampler can also improve the baselines, especially for SSv2 dataset. However, our single temporal selector (TS) still outperforms it on two datasets. Further, when our complete sampler is adopted (TS + SA), we surpass its performance with a significant margin. The above results prove that (1) The spatial sampling is necessary and beneficial (2) \textit{\textbf{Directly applying current methods in general action recognition is not so ideal, while learning a specific sampling strategies for few-shot action recognition is obviously a better solution}}.

\subsection{Study on task-adaptive learner} 
\begin{figure}[ht]
    \centering
    \includegraphics[width=0.8\linewidth]{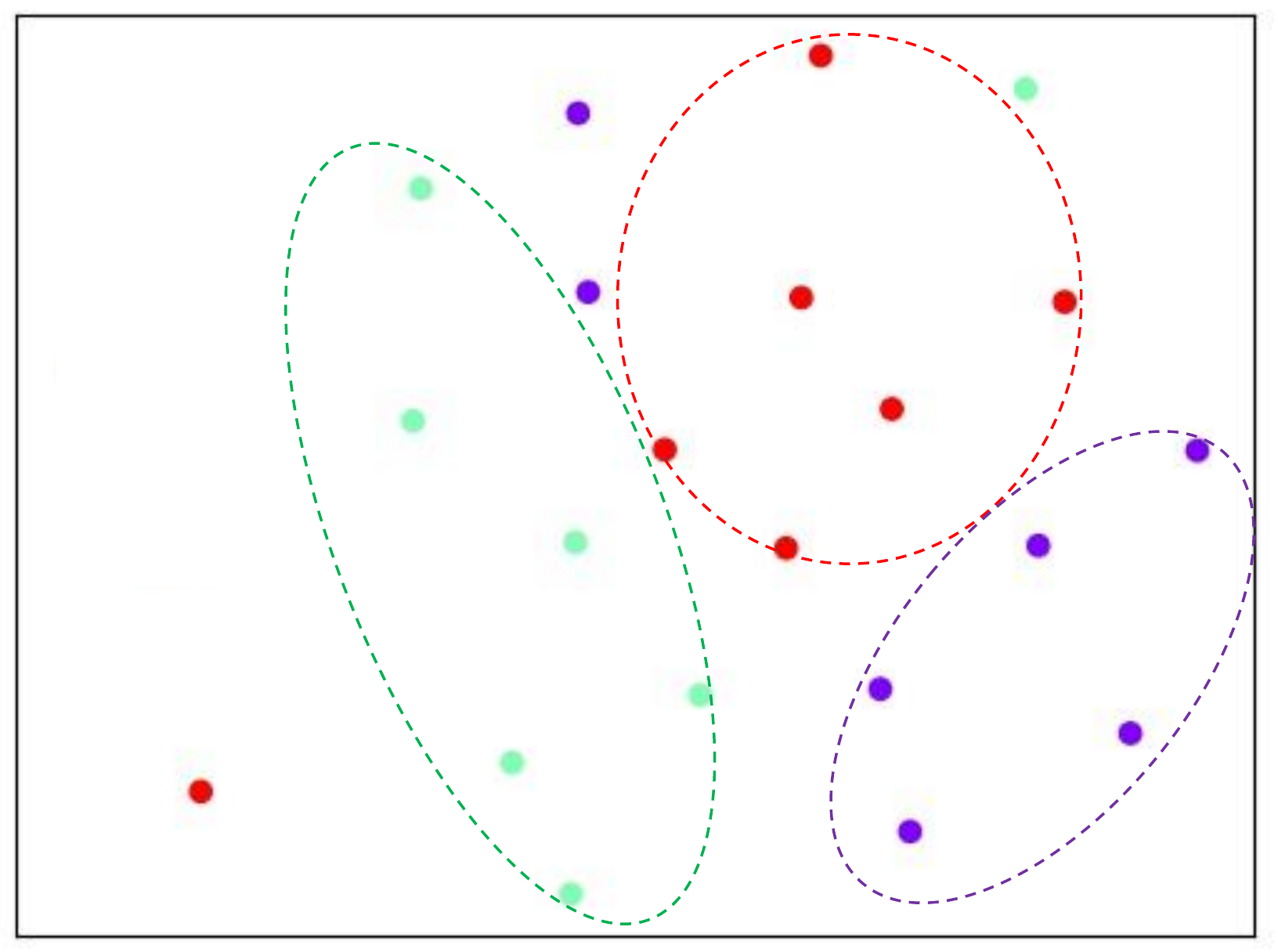}
    %\vspace{-2em}
    \caption{t-SNE visualization of task-specific weights. Different tasks are indicated by different colors.}
    \label{fig:task_tsne}
    %\vspace{-1em}
\end{figure}

To gain further insight into our task-adaptive learner, we visualize the generated parameters for different tasks with t-SNE. Specifically, we randomly select three different tasks and sample 7 task summary features by re-parameterization trick for each task. Then, we can obtain 7 task-specific parameters for $w_2\in \mathbb{R}^{d\times 1}$ by our generator. Finally, we visualize all generated weights of these three tasks with t-SNE. Results are presented in \autoref{fig:task_tsne}. It can be observed that task-specific weights of the same task are closed in feature space, while weights of different tasks own different distributions. It demonstrates that our task-adaptive learner can well adjust the sampling strategies by generating task-specific weights of layers in our sampler network.

\subsection{Details of crop-based spatial sampling} As we have discussed in our paper, we also test another popular crop-based solution for spatial sampling. In this part, we give more details about its implementation. Specifically, a predictor is utilized to predict the center coordinates $(x_c,y_c)$ of a sub-region. Given the pre-defined size $P$ of the sub-region, its spatial patch can be located in the original image. Then, we follow the interpolation-based paradigm~\cite{adafocusv2} to crop this sub-region from the original image in a differential way. In this way, each pixel value in this sub-region can be obtained via bilinear interpolation from four adjacent pixels in the original image.

\begin{table}[ht]
    \centering
    \begin{tabular}{c|c|c}
    \hline
    Models               & UCF101 & SSv2 \\ \hline
    ProtoNet             &   78.7     &  39.3     \\ \hline
    ProtoNet + MGSampler~\cite{mgsampler} &   80.4    &   40.8   \\ \hline
    ProtoNet + Our TS      &    81.1    & 41.3     \\ \hline
    ProtoNet + Our sampler      &   \textbf{82.5}     &   \textbf{41.8}   \\
    \hline
    \end{tabular}
    \caption{Comparison with existed frame sampler in general action recognition. Reported under 1-shot.}
    \label{tab:general}
\end{table}

% \begin{figure*}[hb]
%     \centering
%     \includegraphics[width=0.8\linewidth]{samples/fig/ucf_vis_supp.png}
%     %\vspace{-2em}
%     \caption{Visualization of sampler on UCF101.
%     }
%     \label{fig:ucf_vis}
%     %\vspace{-1em}
% \end{figure*}

\section{Visualization}
\begin{figure*}[ht]
    \centering
    \includegraphics[width=0.95\linewidth]{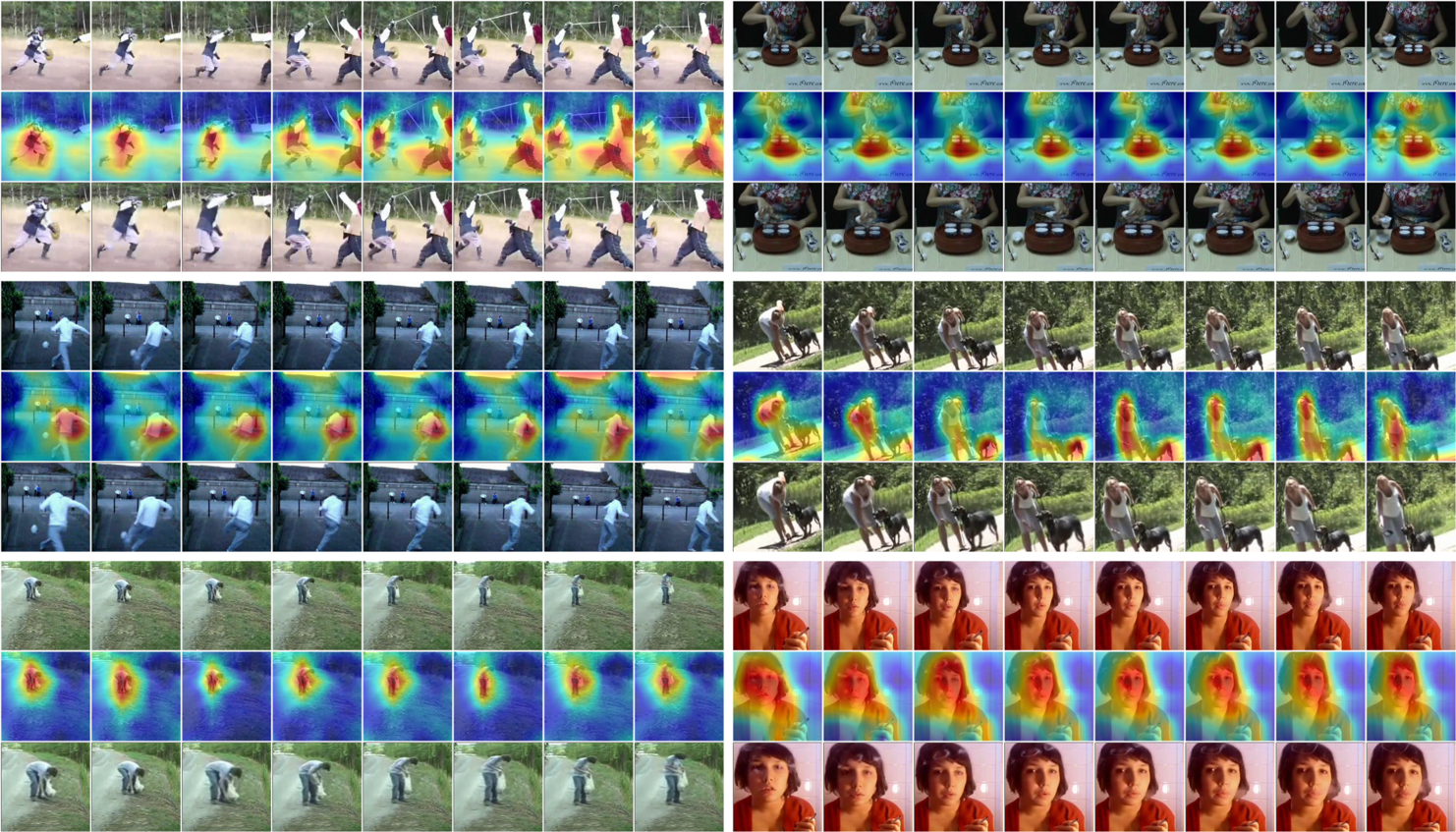}
    %\vspace{-2em}
    \caption{Visualization of sampler on HMDB51.
    }
    \label{fig:hmdb_vis}
    %\vspace{-1em}
\end{figure*}

\begin{figure*}[ht]
    \centering
    \includegraphics[width=0.95\linewidth]{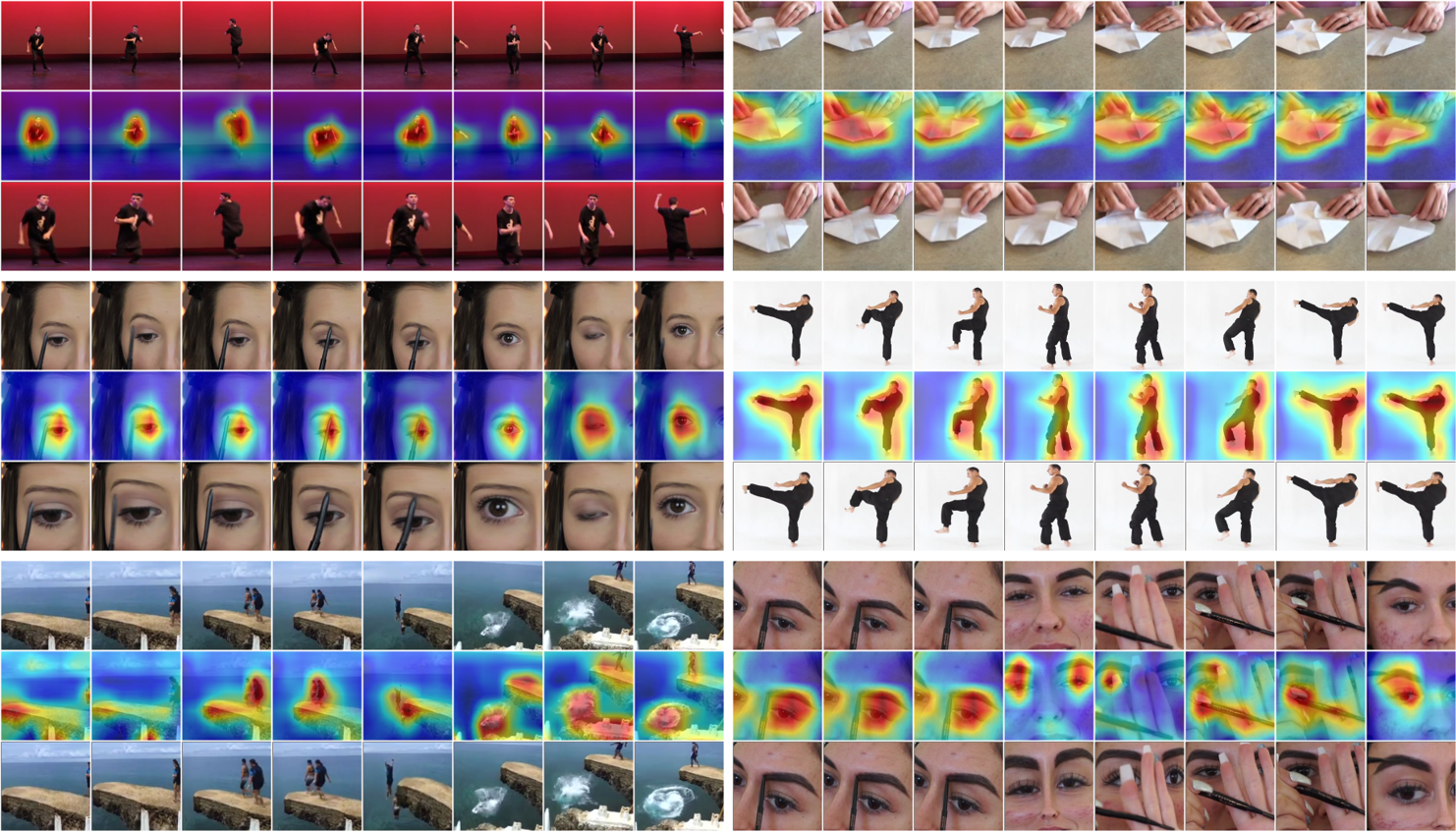}
    %\vspace{-2em}
    \caption{Visualization of sampler on Kinetics.
    }
    \label{fig:kin_vis}
    %\vspace{-1em}
\end{figure*}

In this supplementary, we further visualize more results of our sampler to validate its effectiveness. Please refer to \autoref{fig:hmdb_vis}, and \autoref{fig:kin_vis} for details.

% \section{Reference}
% [1] MGSampler: An Explainable Sampling Strategy for Video Action Recognition, ICCV, 2021

% \noindent
% [2] AdaFocus V2: End-to-End Training of Spatial Dynamic Networks for Video Recognition, arxiv, 2022

\end{document}